\documentclass{article}

\usepackage{microtype}
\usepackage{graphicx}
\usepackage{subfigure}
\usepackage{booktabs} % for professional tables

\usepackage{hyperref}

\usepackage[accepted]{icml2023}

\usepackage{amsmath}
\usepackage{amssymb}
\usepackage{mathtools}
\usepackage{amsthm}

\usepackage[capitalize,noabbrev]{cleveref}

\theoremstyle{plain}

\theoremstyle{definition}

\theoremstyle{remark}

\usepackage[textsize=tiny]{todonotes}

\icmltitlerunning{FACADE: A Framework for Adversarial Circuit Anomaly Detection and
Evaluation}

\begin{document}

\twocolumn[
\icmltitle{FACADE: A Framework for Adversarial Circuit Anomaly Detection and\\Evaluation}

% It is OKAY to include author information, even for blind
% submissions: the style file will automatically remove it for you
% unless you've provided the [accepted] option to the icml2023
% package.

% List of affiliations: The first argument should be a (short)
% identifier you will use later to specify author affiliations
% Academic affiliations should list Department, University, City, Region, Country
% Industry affiliations should list Company, City, Region, Country

% You can specify symbols, otherwise they are numbered in order.
% Ideally, you should not use this facility. Affiliations will be numbered
% in order of appearance and this is the preferred way.
\icmlsetsymbol{equal}{*}
\begin{icmlauthorlist}
\icmlauthor{Dhruv Pai}{equal,cs}
\icmlauthor{Andres Carranza}{equal,cs}
\icmlauthor{Rylan Schaeffer}{equal,cs}
\icmlauthor{Arnuv  Tandon}{equal,cs}
\icmlauthor{Sanmi Koyejo}{cs}
%\icmlauthor{}{sch}
%\icmlauthor{}{sch}
%\icmlauthor{}{sch}
\end{icmlauthorlist}

\icmlaffiliation{cs}{Computer Science, Stanford University}

\icmlcorrespondingauthor{Rylan Schaeffer}{rschaef@cs.stanford.edu}

% You may provide any keywords that you
% find helpful for describing your paper; these are used to populate
% the "keywords" metadata in the PDF but will not be shown in the document
\icmlkeywords{Machine Learning, ICML}

\vskip 0.3in
]

% this must go after the closing bracket ] following \twocolumn[ ...

% This command actually creates the footnote in the first column
% listing the affiliations and the copyright notice.
% The command takes one argument, which is text to display at the start of the footnote.
% The \icmlEqualContribution command is standard text for equal contribution.
% Remove it (just {}) if you do not need this facility.

%\printAffiliationsAndNotice{}  % leave blank if no need to mention equal contribution
\printAffiliationsAndNotice{\icmlEqualContribution} % otherwise use the standard text.

\begin{abstract}
We present FACADE, a novel probabilistic and geometric framework designed for unsupervised mechanistic anomaly detection in deep neural networks. Its primary goal is advancing the understanding and mitigation of adversarial attacks. FACADE aims to generate probabilistic distributions over circuits, which provide critical insights to their contribution to changes in the manifold properties of pseudo-classes, or high-dimensional modes in activation space, yielding a powerful tool for uncovering and combating adversarial attacks. Our approach seeks to improve model robustness, enhance scalable model oversight, and demonstrates promising applications in real-world deployment settings. 
\end{abstract}

\section{Introduction}
\label{sec:intro}

In recent years, the field of machine learning has witnessed significant advancements propelled by improvements in learning algorithms and increased access to computational resources. These advancements have led to the development of larger and more capable models, offering remarkable performance on various tasks. However, as models grow in size and complexity, their interpretability diminishes \cite{10114634}.  The sheer scale of modern deep learning models renders traditional methods of interpretation, such as feature importance or attribution, inadequate. The opacity of these models hinders our ability to understand the reasoning behind their predictions and opens the door to potential adversarial attacks.

Moreover, complex models possess numerous axes along which adversarial attacks can be targeted, making them more vulnerable. Adversarial attacks exploit small, carefully crafted perturbations to inputs that can mislead models into making incorrect predictions \cite{https://doi.org/10.1049/cit2.12028}. With the increase in model complexity, the space of potential adversarial perturbations expands exponentially, making detection and mitigation increasingly difficult.

Furthermore, as AI capabilities continue to grow, and as the freedoms afforded to such models continue to expand, the implications of a hijacked model pose substantial risks to society, for example through damage to critical systems or infrastructure. Robust mechanisms are needed to detect and prevent the misuse of models, especially as they become more powerful and potentially capable of autonomously evolving their behavior. An unsupervised framework for detecting anomalous behavior in models can serve as a crucial component in safeguarding against such risks.

In this paper, we propose mechanistic anomaly detection via probabilistic models for circuit mechanisms within models as a scalable method for model oversight. Our novel circuit-based framework aims to elucidate complex mechanistic pathways relevant to robustness and does so in an unsupervised fashion without any priors as to the nature of an adversarial attack. Our method develops probabilistic models operating in the geometry of neural activation space that facilitates the detection of deviations from the expected behavior, thereby enabling the identification of anomalous model outputs or adversarial attacks%We believe that this framework represents a significant step towards enhancing model interpretability, and strengthening security and safety in the field of (adversarial) machine learning.
\footnote{To understand how our proposal relates to emerging directions in adversarial machine learning, see \citet{carranza2023dam}.}.

\section{Mechanistic Anomaly Detection}
\label{sec:mad}

\subsection{Activations}
Previously, a detailed analysis of flows in activation space proved computationally intensive and opaque. Insofar as neural networks apply a series of nonlinear geometric transformations to high-dimensional data manifolds, the propagation of data points through these transformations is computationally irreducible and largely uninterpretable directly \cite{cohen2020separability}. However, understanding the propagation of data points in the high-dimensional activation space has profound implications for the reliability and security of our models, and this understanding may hold the key to investigating adversarial robustness.

Adversarial examples have been demonstrated to exploit the complex, and often poorly understood, geometry of the decision boundaries within this high-dimensional space \cite{gebhart2019characterizing}. Thus, developing methods capable of elucidating these decision boundary structures and understanding the geometry of high-dimensional modes in activation space would lead to improved adversarial robustness. Prior literature and preliminary experiments demonstrate that neural networks in activation space learn pseudo-classes:  intermediate groupings of features learned by the model that resemble high-density modes \cite{gebhart2019characterizing}. An understanding of the distribution, composition, and shape of pseudo-classes within a network offers a lens into the mechanistic behavior of the model. 

\subsection{Circuit Mechanisms}
As defined by \citet{wang2022interpretability}, a circuit is a subgraph of a neural network's overall computational graph. Small circuits have recently been investigated for their role in interpretability, in particular identifying circuits corresponding to certain visually meaningful properties of an image, e.g., orientation, curve-detection, color-detection circuits \cite{olah2020zoom}. Circuits have been demonstrated as a valuable intermediate between single-neuron and whole-model holistic interpretability, as they are well-conserved across mechanistically similar models and provide valuable insights into model behavior for a wide variety of architectures and datasets \cite{elhage2021mathematical}. 
\begin{figure}[H]
    \centering
    \includegraphics[scale=0.29]{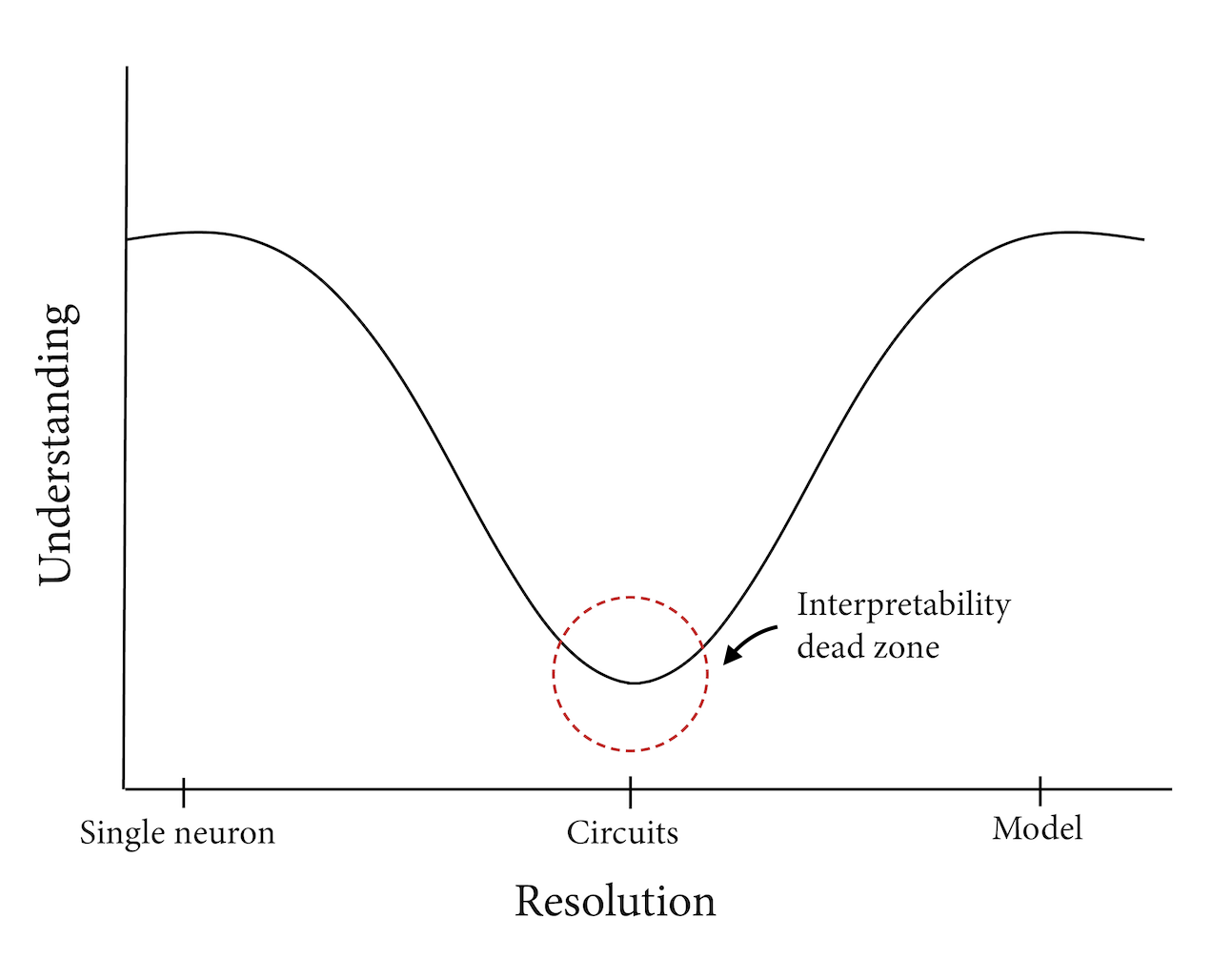}
    \caption{The intermediate scale of interpretability is simultaneously the most poorly understood and the most complex, yet could offer the greatest insights into MAD.}
    \label{fig:my_label}
\end{figure}
Problematically, circuit interpretability has focused overwhelmingly on uncovering circuits responsible for specific adversarially or visually meaningful features specified \textit{a posteriori} \cite{olah2020zoom, wang2022interpretability, conmy2023automated}. Such an approach requires a specification of the features used in an adversarial attack before circuit interpretability is applied, which is rarely the case in a real-world deployment setting \cite{carlini2019evaluating}. Supervised circuit interpretability also extrapolates poorly across mechanistically distinct models. We therefore motivate an unsupervised circuit interpretability approach with the promise of revealing novel insights into mechanistic anomalies, while improving model invariance, computational efficiency, and scalable model oversight. 

\subsection{Blue Sky MAD Approach}
To identify adversarial examples, and their corresponding circuits through mechanistic anomaly detection, we propose a novel probabilistic, geometric framework for creating unsupervised distributions over circuits in a deep neural network titled FACADE: A Framework for Adversarial Circuit Anomaly Detection and Evaluation.
Specifically, we envision a four-step approach, where the hyperparameter $\lambda$ can be interpreted as setting the resolution of circuit analysis.  
\begin{enumerate}
    \item Utilize the probabilistic Dirichlet Process Mixture model \cite{blei2006variational,kulis2011revisiting} for unsupervised clustering (DP-Means) to identify "pseudoclass" modes in intermediate activation space for a given density threshold $\lambda$ \cite{dinari2022revisiting}
    \item Elucidate circuits responsible for pseudoclass formation and propagation through causal discovery and Automatic Circuit DisCovery (ACDC) \cite{nauta2019causal} \cite{conmy2023automated}
    \item Determine manifold and kernel density properties of pseudoclass propagation through circuits and in relation to final classes through mean-field theoretic approximation \cite{cohen2020separability}
    \item Generate a distribution over circuits as they contribute to changes in manifold properties of pseudoclasses as they propagate through the network, e.g. effective reduction in radius or dimension
\end{enumerate}
Repeating the above algorithm for a sweep of $\lambda$ values allows for circuit distribution evaluation across a variety of features and mechanistic pathways. By analyzing anomalous circuits in the distribution or employing FACADE to prune circuits, we envision significant gains in adversarial robustness. Adversarial circuits, identified as probabilistic outliers in geometric transformations, would stand out on FACADE distributions and could easily be reverse-engineered to derive how adversarially-susceptible pseudoclasses can be made more robust by surgical tuning of weights. It is worth noting that FACADE relies on sufficiently many training examples to capture meaningful activation flows in an unsupervised fashion. However, if this condition is met, at test-time FACADE distributions with a simple probabilistic thresholding approach can identify and prevent mechanistic anomalies and adversarial attacks autonomously. 
\clearpage

\bibliographystyle{icml2023}
\bibliography{references}

\end{document}